\documentclass[11pt,a4paper]{article}
\usepackage[hyperref]{emnlp2018}
\usepackage{times}
\usepackage{latexsym}
\usepackage{graphicx}
\usepackage{url}
\usepackage{multirow}
\usepackage{epsfig}
\usepackage{CJKutf8}
\PassOptionsToPackage{normalem}{ulem}
\usepackage{CJKulem}
\usepackage{xcolor}
\usepackage{amsmath}
\usepackage{bm}
\usepackage[group-separator={,},group-minimum-digits={3}]{siunitx}
\usepackage{subfigure}
\usepackage{epsfig}
\usepackage{amssymb}
\usepackage{multirow}

\aclfinalcopy 

\newcommand*{\affaddr}[1]{#1} 
\newcommand*{\affmark}[1][*]{\textsuperscript{#1}}
\newcommand*{\email}[1]{\texttt{#1}}

\newcommand{\fancyhl}[3][red]{\colorlet{#1#2}{#1!#2}\bgroup\markoverwith{\textcolor{#1#2}{\rule[-.5ex]{0.4pt}{2.5ex}}}\ULon{#3}}
\newcommand{\fancyhc}[3][green]{\colorlet{#1#2}{#1!#2}\bgroup\markoverwith{\textcolor{#1#2}{\rule[-.5ex]{0.4pt}{2.5ex}}}\ULon{#3}}

\title{Evaluating Semantic Rationality of a Sentence: A Sememe-Word-Matching Neural Network based on HowNet}

\author{Shu Liu\affmark[1,2], Jingjing Xu\affmark[2], Xuancheng Ren\affmark[2], Xu Sun\affmark[1,2]\\
\affaddr{\affmark[1]Deep Learning Lab, Beijing Institute of Big Data Research, Peking University}\\
\affaddr{\affmark[2]MOE Key Lab of Computational Linguistics, School of EECS, Peking University}\\
\email{\{shuliu123, jingjingxu, renxc, xusun\}@pku.edu.cn}\\
}

\date{}

\begin{document}
\maketitle
\begin{abstract}

Automatic evaluation of semantic rationality is an important yet challenging task, and current automatic techniques cannot well identify whether a sentence is semantically rational. The methods based on the language model do not measure the sentence by rationality but by commonness. The methods based on the similarity with human written sentences will fail if human-written references are not available. In this paper, we propose a novel model called Sememe-Word-Matching Neural Network (SWM-NN) to tackle semantic rationality evaluation by taking advantage of sememe knowledge base HowNet. The advantage is that our model can utilize a proper combination of sememes to represent the fine-grained semantic meanings of a word within the specific contexts. We use the fine-grained semantic representation to help the model learn the semantic dependency among words. To evaluate the effectiveness of the proposed model, we build a large-scale rationality evaluation dataset. Experimental results on this dataset show that the proposed model outperforms the competitive baselines with a 5.4\% improvement in accuracy. 

\end{abstract}

\section{Introduction}

Recently, tasks involving natural language generation has been attracting heated attention. However, it remains a problem how to measure the quality of the generated sentences most reasonably and efficiently. Chomsky's famous sentence, ``colorless green ideas sleep furiously''~\cite{chomsky1956three}, is correct in syntax but irrational in semantics. It is important to measure the quality of these sentences. Conventional methods involve human judgments of different quality metrics. However, it is both labor-intensive and time-consuming to conduct a human evaluation, which inevitably involves the problem of the fairness of the metrics and the agreement among the annotators. In this paper, we explore an important but challenging problem: how to automatically identify whether a sentence is semantically rational. Based on this problem, we propose an important task: Sentence Semantic Rationality Detection (SSRD), which aims to identify whether the sentence is rational in semantics. The task can benefit many natural language processing applications that require the evaluation of the rationality and can also provide insights to resolve the irrationality in the generated sentences. 

There exist some automatic methods to evaluate the quality of a sentence. Some are based on language models, some are based on similarity with human-written sentences, and others are the machine learning models based on statistical features. The drawback of language models ~\cite{kneser1995improved,chen1999empirical} lies in that these models do not measure the sentence by rationality but by commonness. The commonness is represented by the probability of a sentence in the space of all possible sentences. Considering that the uncommon sentences are not always irrational, this approach is not a suitable solution. For similarity-based methods ~\cite{papineni2002bleu,lin2004rouge,xu2016optimizing}, the evaluation criterion is the $N$-gram matching numbers between the generated sentences and the reference sentences. If gold answers are given, it is indeed a valuable evaluation method. However, the main problem lies in that in many NLP tasks, human-written references are not available, which leads to the failure of this method. For some statistical feature-based methods such as decision tree~\cite{eneva2001learning}, they only use the statistical information of the sentence. However, it is also essential to use the semantic information of the sentences when evaluating their quality.

The main difficulty in the evaluation of semantic rationality is that it requires systems with high ability to understand selectional restrictions. In linguistics, \textit{selection} denotes the ability of predicates to determine the semantic content of their arguments. Predicates select their arguments, which means that they limit the semantic content of their arguments. The following example illustrates the concept of selection. For a sentence ``The building is wilting'', the argument ``the building'' violates the selectional restrictions of the predicate, ``is wilting''. To address this problem, we propose to take advantage of the sememe knowledge which gives a more detailed semantic information of the word. Using this sort of knowledge, a model would learn the selectional restrictions between words better.

Words can be represented with semantic sub-units from a finite set of limited size. For example, the word ``lover'' can be approximately represented as ``\{Human $|$ Friend $|$ Love $|$ Desired\}''. Linguists define \textit{sememes} as semantic sub-units of human languages ~\cite{bloomfield1926set} that express semantic meanings of concepts. One of the most well-known sememe knowledge bases is HowNet ~\cite{zhendong2006hownet}. HowNet has been widely used in various Chinese NLP tasks, such as word sense disambiguation ~\cite{duan2007word}, named entity recognition ~\cite{10.1007/978-3-319-50496-4_38}, word representation ~\cite{niu2017improved}, ~\newcite{zeng2018chinese} proposed to expand the Linguistic Inquiry and Word Count ~\cite{pennebaker2001linguistic} lexicons based on word sememes, and ~\newcite{li2018sememe} proposed a task called sememe prediction to learn semantic knowledge from unstructured textual Wiki descriptions.

In this work, we address the task of automatic semantic rationality evaluation by using the semantic information expressed by sememes. We design a novel model by combining word-level information with sememe-level semantic information to determine whether the sememes of the words are compatible so that the sentence does not violate common perception. We divide our model into two parts: a word-level part and a sememe-level part. First, the word-level part gets the context for each word. Next, we use the context of each word to select its proper sememe-level information. Finally, we use information of both levels to predict the semantic rationality of a sentence.

Our main contributions are listed as follows:

\begin{itemize}
\item We propose the task of automatically detecting sentence semantic rationality and we build a new and large-scale dataset for this task. 

\item We propose a novel model called SWM-NN that combines sentence information with its sememe information given by Chinese knowledge base HowNet. Experimental results show that the proposed method outperforms the baselines. 
\end{itemize}

\section{Proposed Method}

\begin{figure}[tb]
	\centering
    \includegraphics[width=1.0\linewidth]{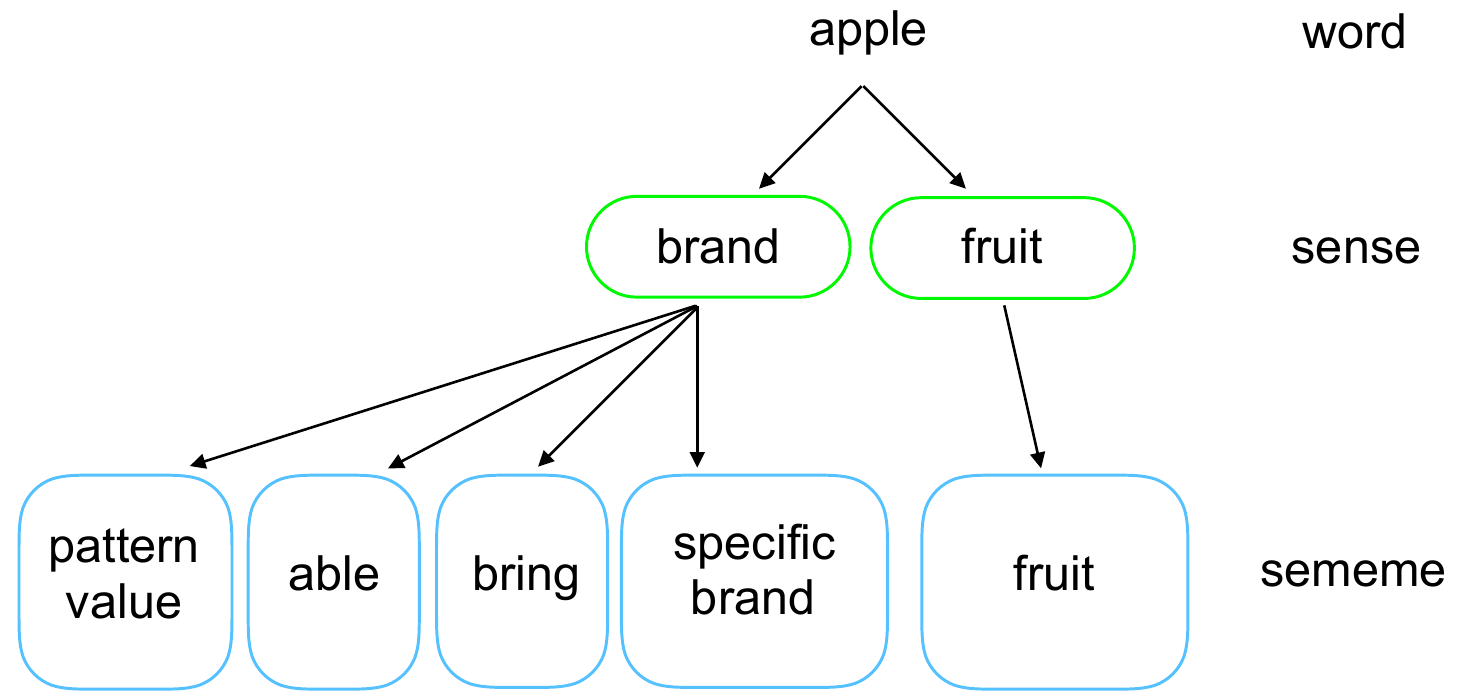}
	\caption{Examples of sememes, senses, and words. We translate them into English.}\label{Fig:HowNet_em}
	\vspace{-0.05in}
\end{figure}

To detect the semantic rationality of the sentence, we should represent the sentence into fine-grained semantic units. We deal with the task of SSRD with the aid of the sentence representation and its semantic representation.

\begin{figure*}[t]
    \vspace*{-1.0cm}
    \hspace*{-0.8cm}
    \includegraphics[width=7.0in]{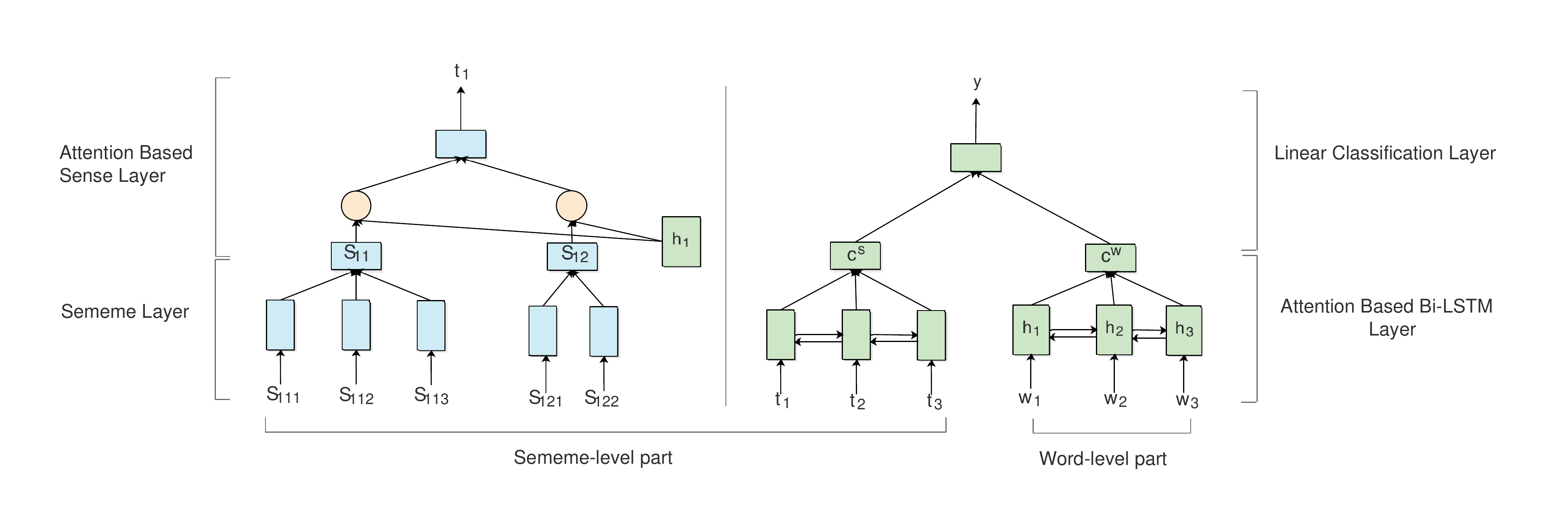}
	\caption{The overview of SWM-NN model. The left part consists of the sememe layer and the sense layer. The right part consists of two layers. One is the attention based LSTM of the sememe-level and the word-level, another is the linear classification layer. In the example, the word $w_1$ has two senses $\bm{s}_{11}$ and $\bm{s}_{12}$. The sense $\bm{s}_{11}$ has three sememes $\bm{s}_{111}$, $\bm{s}_{112}$, $\bm{s}_{113}$. The sense $\bm{s}_{12}$ has two sememes $\bm{s}_{121}$, $\bm{s}_{122}$.} \label{Fig:model_fig}
	\hspace{-0.8in}
    \vspace{-0.2in}
\end{figure*}

Based on this motivation, we proposed an SWM-NN model (see Figure~\ref{Fig:model_fig}). This model can make use of HowNet, which is a well-known Chinese semantic knowledge base. The overall architecture of SWM-NN consists of several parts: a word-level attention LSTM, a matching mechanism between the word-level and the sememe-level part, and a sememe-level attention LSTM. We first introduce the structures of HowNet, and then we describe the details of different components in the following sections.

\subsection{Sememes, Senses and Words in HowNet}
HowNet annotates precise \textbf{senses} to each word, and for each \textbf{sense}, HowNet annotates the significant of parts and attributes represented by \textbf{sememes}. Figure~\ref{Fig:HowNet_em} shows the sememe annotations of the word ``apple''. We translate this part of HowNet into English. The word ``apple'' actually has two main \textbf{senses}: one is a sort of juicy fruit ``fruit'', and the other is a famous computer brand ``brand''. The latter \textbf{sense} ``Apple brand'' indicates a computer brand, and thus has \textbf{sememes} ``computer'', ``bring'', ``Special Brand''. 

We introduce the notations used in the following sections as follows. Given a sentence $\bm{s}$ consisting of a sequence of words $\{\bm{d}_1, \bm{d}_2, \cdots, \bm{d}_n\}$, we embed the one-hot representation of the $i$-th word $\bm{d}_i$ to a dense vector $\bm{w}_i$ through a word embedding matrix. For the $i$-th word $\bm{d}_i$, there can be multiple senses $\bm{s}_j^{(\bm{d}_i)}$ in HowNet. Each sense $\bm{s}_j^{(\bm{d}_i)}$ consists of several sememe words $\overline{\bm{d}}_k^{(s_j)}$ in HowNet. The one-hot representation of the sememe word $\overline{\bm{d}}$ is embedded to a dense vector $\bm{x}$ through a sememe embedding matrix.

\subsection{Word-Level Attention LSTM}

To detect the rationality using sentence information, we use a Bi-LSTM encoder with local attention in the word-level part. We first compute the context output $\bm{o}^w = \{\bm{o}_1^w, \bm{o}_2^w, \cdots, \bm{o}_L^w \}$ from the source sentence $\bm{w} =\{\bm{w}_1, \bm{w}_2, \cdots, \bm{w}_L\}$:
\begin{align}\label{wordencoder}
\overrightarrow{\bm{o}}_i^{w}, \overrightarrow{\bm{h}}_i^{w} &= {\rm LSTM}_{word}(\bm{w}_i,\overrightarrow{\bm{h}}_{i-1}^{w}) \\
\overleftarrow{\bm{o}}_i^{w}, \overleftarrow{\bm{h}}_i^{w} &= {\rm LSTM}_{word}(\bm{w}_i,\overleftarrow{\bm{h}}_{i+1}^{w}) \\
\bm{h}_i^{w} &= [\overrightarrow{\bm{h}}_i^{w}; \overleftarrow{\bm{h}}_i^{w}] \\
\bm{o}_i^{w}&=[\overrightarrow{\bm{o}}_i^{w}; \overleftarrow{\bm{o}}_i^{w}]
\end{align}
where $L$ is the number of words in the source sentence. Then, we use the context output $\bm{o}^w=\{\bm{o}_1^w, \bm{o}_2^w, \cdots, \bm{o}_L^w \}$ to compute an attention vector $\bm{\alpha}^{w}=\{\alpha_1^{w}, \alpha_2^{w}, \cdots, \alpha_L^{w}\}$. Finally, we use the context output $\bm{o}^w$ and the attention vector $\bm{\alpha}^{w}$ to compute a word-level representation of the sentence $\bm{c}^w$. The calculation formulas are as follows:
\begin{align}\label{attentioncontext}
\bm{u}_i^{w} &= \tanh (\bm{W}_w \bm{o}_i^{w}+\bm{b}_w)  \\
\alpha_i^{w} &= \frac{\exp \left((\bm{u}_i^w)^T \bm{u}_w\right)}{\sum_j \exp \left((\bm{u}_j^w)^T \bm{u}_w\right)} \\
\bm{c}^w &= \sum_i \alpha_i^w \bm{o}_i^w
\end{align}
where $\bm{W}_w$ and $\bm{b}_w$ are weight matrix and bias vector, respectively. $\bm{u}_w$ is a randomly initialized vector, which can be learned at the training stage. The attention mechanism is proposed in~\cite{bahdanau2014neural}, which gives higher weights to certain features that allow better prediction. Through training, the certain feature is likely to be the word that destructs the rationality of the sentence in semantics.

\subsection{Matching Mechanism Layer}
In sememe-level part, we average the sememe embeddings to represent each sense of the word $\bm{d}$ at first:
\begin{equation}\label{sem_ave}
\bm{s}_j^{(\bm{d})} = \frac{1}{m_j^{(\bm{d})}}\sum\limits_{k} \bm{x}_k^{(\bm{s}_j)}
\end{equation}
where $\bm{s}_j^{(\bm{d})}$ stands for the $j$-th sense embedding of the word $\bm{d}$. $m_j^{(\bm{d})}$, $\bm{x}_k^{(\bm{s}_j)}$ stands for the number of sememes and the $k$-th sememe embedding belonging to the $j$-th sense of $\bm{d}$ (i.e. $\bm{s}_j^{(\bm{d})}$), respectively. Hence, given a word $\bm{d}_i$, we can get the sense embedding matrix of $\bm{d}_i$, referred to as $
\bm{S^{(d_i)}}=[\bm{s}_1^{(\bm{d_i})}, \bm{s}_2^{(\bm{d_i})}, \cdots, \bm{s}_{n_i}^{(\bm{d_i})}]$, where $n_i$ stands for the number of senses belong to $\bm{d_i}$.

To match the appropriate senses and sememes to each word given a specific sentence, we add a matching mechanism that is based on global attention. Since the output of word-level LSTM $\bm{o}_i^{w}$ can be viewed as the contextual representation. For each word $\bm{d}_i$, we have the output state $\bm{o}_i^{w}$ in word-level LSTM and its sense embedding matrix $\bm{S^{(\bm{d}_i)}}=[\bm{s}_1^{(\bm{d}_i)}, \bm{s}_2^{(\bm{d}_i)}, \cdots, \bm{s}_{n}^{(\bm{d}_i)}]$. 

We compute the sememe-level representation $\bm{t}_i$ of the word $\bm{d}_i$ as follows:
\begin{align}
\beta_{j} &= \frac{\exp \left(g(\bm{o}_i^{w}, \bm{s}_j^{(\bm{d}_i)})\right)}{\sum_k \exp \left(g(\bm{o}_i^{w}, \bm{s}_k^{(\bm{d}_i)})\right)}  \\
\bm{t}_i &= \sum \limits_{j} \beta_j \bm{s}_j^{(\bm{d}_i)}
\end{align}
Here the score function $g$ is computed as follows:
\begin{equation}
g(\bm{o}_i^{w}, \bm{s}_j^{(\bm{d}_i)}) = \tanh(\bm{W}_x\bm{o}_i^{w})\odot \tanh(\bm{W}_y\bm{s}_j^{(\bm{d}_i)}) \\
\end{equation}
where $\bm{W}_x$ and $\bm{W}_y$ are model parameters, which can be learned at the training stage.

\subsection{Sememe-level Attention LSTM}
For each sentence $s=\{\bm{d}_1, \bm{d}_2, \cdots, \bm{d}_n\}$, we can get its sememe-level sequences $\{\bm{t}_1, \bm{t}_2, \cdots, \bm{t}_n\}$ based on the computation mentioned above. We use a sememe-level attention LSTM, which is similar to the word-level attention LSTM, to get the sememe-level representation of the sentence $\bm{c}^s$.

\subsection{Combining Information from the two parts}
We take the vector $\bm{c}^w$, $\bm{c}^s$ as input, and use a $softmax$ layer to predict the probability of the semantic rationality. 

For each sentence, the probability distribution of the label is
\begin{equation}\label{softmaxpre}
p(\hat{y}|\bm{w}, \bm{s}) = softmax(\bm{W}^w \bm{c}^w + \bm{W}^s \bm{c}^s + \bm{b})
\end{equation}
where $\hat{y}$ is the label of whether the sentence is rational in semantics. $\bm{w}$, $\bm{s}$ is the word sequence and its sememe sequence respectively.

$\bm{\theta}$ is the model parameter and $y$ is the ground-truth label of the sentence, then the cross entropy loss is
\begin{equation}\label{crossloss}
\mathcal{L}(\theta) = -y \log p(y|\bm{w}, \bm{s}, \bm{\theta})
\end{equation}

\section{Experiments}

In this section, we evaluate our model on the dataset we build for the SSRD task. Firstly, we introduce the dataset and the experimental details. Then, we compare our model with baselines. Finally, we provide the analysis and the discussion of the experimental results.

\subsection{Dataset}

We create the dataset by collecting Chinese Word Segmentation and Part-of-Speech Tagging corpus from China National Language Committee\footnote{\url{http://www.aihanyu.org/cncorpus/}}. Then we divide the dataset into training, validation, and test set. To create the sentences that lack rationality in semantics (i.e. the negative samples), we randomly do one of the following operations on every sentence in each set: 

\begin{enumerate}
\item Replace word with the same POS randomly.
\item Reverse the position of two words of the same POS randomly. 
\item Generate the sentence using the 5-gram language model.
\end{enumerate}

We randomly choose 300 true sentences and 300 permuted sentences (including 100 permuted sentences for each operation), then we ask some human annotators (highly educated native speakers) to label whether the 600 sentences are fluent while asking some other human annotators to label whether they are rational in semantics. The results are shown in Table \ref{permuted_sentences}. It shows that compared to the true sentences, the permuted sentences which are created by the first two operations is only a little different in fluency starkly different in rationality. The third method, however, is starkly different in both fluency and semantic-rationality compared to the true sentences. To make the dataset better match the purpose of our task, we only use one of the first two generate methods (i.e. Replace two words with the same POS randomly or Reverse the position of two words of the same POS randomly). In order to ensure the irrationality of the permuted sentences, we permuted sentences of which length is more than $8$ and we didn't replace punctuation of sentence.

\begin{table}[tb]
		\centering
        \setlength{\tabcolsep}{1mm}{
		\begin{tabular}{p{100pt} c c}
		\hline
		\multicolumn{1}{l}{\textbf{Dataset}} &
		\multicolumn{1}{c}{\textbf{Fluency}} &  
		\multicolumn{1}{c}{\textbf{Rationality}}   \\ \hline
		300 true sentences & 100\% & 99.3\% \\  \hline
		100 permuted sentences for operation1& 64\% & 3\% \\ \hline
		100 permuted sentences for operation2& 70\% & 6\% \\ \hline
        100 permuted sentences for operation3& 3\% & 1\% \\ \hline
		\end{tabular}}
		\caption{Fluency and Semantic-Rationality of the true sentences and permuted sentences labeled by some human annotators. }
		\label{permuted_sentences}
	\end{table}

The details of each set are shown in Table \ref{tab_dataset_numbers}.

\subsection{Experimental Details}

We use accuracy as our evaluation metric instead of the F-score, precision, and recall because the positive and negative examples in our dataset are balanced. As the words and the sememes are different in meaning, we do not share their vocabulary. We build up vocabularies for words and sememes with the size of \num{50000} and \num{20000}, respectively. 
Some words are not annotated and thus have no sememes in HowNet. We simply use the word itself as the sememe.

We use the same dimension of 128 for word embeddings and sememe embeddings, and they are randomly initialized and can be learned during training. Adam optimizer~\cite{KingmaBa2014} is used to minimize cross entropy loss function. We apply dropout regularization~\cite{dropout} to avoid overfitting and clip the gradients~\cite{gradientclip} to the maximum norm of 5.0. During training, we train the model for 20 epochs and monitor its performance on the validation set after every 200 updates. Once training is finished, we select the model with the highest accuracy on the validation set as our final model and evaluate its performance on the test set.

\begin{table}[tb]
		\centering
        \setlength{\tabcolsep}{1mm}{
		\begin{tabular}{l  c  c  c}
		\hline
		\multicolumn{1}{l}{\textbf{Dataset}} &
		\multicolumn{1}{c}{\#\textbf{Total}} & 
		\multicolumn{1}{c}{\#\textbf{Positive}} &  
		\multicolumn{1}{c}{\#\textbf{Negative}}   \\ \hline
		Training set & \num{160000} & \num{80000} & \num{80000} \\ 
		Validation set & \num{20000} & \num{10000} & \num{10000} \\ 
		Test set & \num{20000} & \num{10000} & \num{10000} \\ \hline
		\end{tabular}}
		\caption{Statistical information of the final dataset. \textbf{Positive} and \textbf{Negative} denote whether the sentence is semantic rational.}
		\label{tab_dataset_numbers}
	\end{table}
    
\subsection{Baseline Models}

\begin{itemize}
\item \noindent\textbf{$N$-gram language model:} 

Here, we use the best performing $N$-gram smoothing methods, the interpolated Kneser-Ney algorithm~\cite{kneser1995improved,chen1999empirical}. The positive sentences in the training set are used to train the model. For detecting rationality, we calculate a threshold based on the validation set that maximizes the accuracy. Then, we predict the test set using the model and the threshold.

\item \noindent\textbf{Traditional machine learning algorithms:} 
 We use various machine learning classifiers to predict the labels based on the tf-idf features of the sentence. We compute the probability distribution of the label by inputting the sentence word sequence and its sememe word sequence to the model respectively. Then we ensemble the probability of both sequences to get the label prediction.

 \item \noindent \textbf{Neural networks models:} We apply two representative neural network models:  Bi-LSTM~\cite{Hochreiter1997Long} and CNN~\cite{cnn}. The neural network is used to learning the vector representation for the word sequence and the sememe sequence, respectively. Then concatenate the both outputs to input a linear classifier. 
 
 \item \noindent \textbf{Human evaluation:} For 500 randomly chosen sentences, we provide human annotators with the true sentence and the permuted sentence. Then we ask them to select a better sentence. The result can be viewed as an upper bound for this task.
 
\end{itemize}

\subsection{Results}

In this subsection, we present the results of evaluation by comparing our proposed method with the baselines. Table~\ref{tab_part1_res} reports experimental results of various models. From the results, we can observe that:
\begin{itemize}
\item The proposed SWM-NN outperforms all the baselines except the human evaluation. Our model uses dual-attention mechanism that consists of local attention in both levels, and a global attention to match the word to its appropriate combination of the sememes. By properly incorporating knowledge in HowNet and information of the source sentence, our model is capable of making more accurate predictions. 
\item We see that the interpolated Kneser-Ney language model get lowest prediction accuracy in the baseline model. It partly verifies our arguments on resolving the task using language models.
\item The traditional machine learning algorithms with sememe information only achieve the accuracy of 60.5\% at best. Neural network based methods perform much better and beat other baselines. This shows that the generalization ability of neural networks is better (the positive sentences and its similar negative sentences only coexist in the same data set). 
However, the neural network with the sememe information given by HowNet only achieves the accuracy of 63.5\% at best. It suggests merely providing the sememes to the models is not sufficient for detecting rationality. Further matching of sememes to check the compatibility of the sememes is crucial to the overall performance. 
\end{itemize}

\begin{table}[tb]
	\centering
	\begin{tabular}{l c l c}
		\hline
		\multicolumn{1}{l}{\textbf{Models}} &
        \multicolumn{1}{l}{\textbf{Accuracy}}  \\ \hline
        Interpolated Kneser-Ney & 53.2\%  \\ \hline
        Random Forest & 60.5\%   \\
        Linear SVM & 58.7\% \\
        SVM & 57.1\%   \\
        Naive Bayes   & 54.6\% \\ \hline
        
        CNN & 62.7\%  \\
        Bi-LSTM & 63.5\%  \\ \hline
        
         \textbf{SWM-NN} & \textbf{68.9\%} \\ \hline
         \textbf{Human Evaluation} & \textbf{94.4\%} \\ \hline
	\end{tabular}
	\caption{Comparison between our proposed model and the baselines on the test set. Our proposed model is denoted as \textbf{SWM-NN}.}\label{tab_part1_res}
\end{table}

\section{Analysis and Discussions}

Here, we perform further analysis on the model, including the ablation study, case study, error analysis, and some further experiment results.
\subsection{Exploration on Internal Structure of the Model}

As shown in Table~\ref{tab_part1_res}, our SWM-NN model outperforms all the baselines. Compared with the baseline neural network model, the proposed model has a dual-attention mechanism, that is, (1) a local attention mechanism in both the word-level and the sememe-level and (2) a global attention mechanism to match information between two levels. In order to explore the impact of internal structure of the model, we remove the components of our model in order. The performance is shown in Table~\ref{tab_part1_res2}.

\begin{itemize}

\item \textbf{w/o Match} means that we do not match the context of the word to its sememe information by the global attention mechanism. For each word $\bm{d}$, we only average all the sememe embedding to get $\bm{t}$ as follows: 
\begin{equation}\label{computetave}
\bm{t} = \sum \limits_{j} \frac{1}{n} \bm{s}_j^{(d)}
\end{equation}
$n$ is the number of senses of this word.

\item \textbf{w/o Dual-attention} means that we do not use dual attention mechanism (i.e. local attention in both levels and global attention between two levels) in the proposed model any more, which is the same as the Bi-LSTM in the baseline models. 

\item \textbf{w/o HowNet} means we do not use the knowledge given by HowNet. It is equivalent to without sememe-level local attention and matching mechanism.

\item \textbf{w/o Word-level part} means that we do not use word-level information any more. For each sentence word, we only average all the sememe embedding and we only use $\bm{c}^s$ to predict label.

\item \textbf{w/o Word-level $\bm{c}^w$} means without word-level representation, that is, we only use $\bm{c}^s$ to predict label. But we still use other structures of SWM-NN. 

\end{itemize}

\begin{table}[t]
	\centering
	\begin{tabular}{@{}l@{}c@{}c@{}}
		\hline
		\multicolumn{1}{l}{\textbf{Models}} &
		\multicolumn{1}{l}{\textbf{Accuracy}} &
        \multicolumn{1}{l}{\textbf{Decline}} \\ \hline
        SWM-NN & 68.9\% & $--$ \\ \hline
		w/o Match & 67.8\% & $\downarrow$ 1.1\% \\
		w/o Dual-attention & 63.5\% & $\downarrow$ 5.4\% \\ 
        w/o HowNet & 67.2\% & $\downarrow$ 1.7\% \\
        w/o Word-level part &  59.3\% & $\downarrow$   9.6\% \\
        w/o Word-level $\bm{c}^w$&   68.1\% & $\downarrow$   0.8\% \\
        \hline
	\end{tabular}
	\caption{Ablation Study.}\label{tab_part1_res2}
\end{table}

From the results shown in Table~\ref{tab_part1_res2}, we can observe that:

\begin{table*}[tb]
	\centering
    \tiny
	\begin{tabular}{@{}l@{}l@{}l@{}}
		\hline
        
        \multicolumn{1}{l}{\textbf{Test Sentence}} &
		\multicolumn{1}{l}{\begin{CJK*}{UTF8}{gbsn}\textit{全总\ 等\ 单位\ 慰问\ 本\ \textbf{教师}\ \textbf{市}}\end{CJK*}}   \\

        \multicolumn{1}{l}{} &
        \multicolumn{1}{l}{The city's Federation of Trade Unions and other units convey greetings to city in our teachers.} \\ 
        
        \multicolumn{1}{l}{\textbf{Word-level attention}} &
		\multicolumn{1}{l}{\begin{CJK*}{UTF8}{gbsn}\textit{\fancyhc{100}{全总}\ \fancyhc{6.9}{等}\ \fancyhc{20.1}{单位}\ \fancyhc{22.2}{慰问}\ \fancyhc{7.9}{本}\ \fancyhc{0.9}{教师}\ \fancyhc{10.9}{市}}\end{CJK*}} \\
        
        \multicolumn{1}{l}{\textbf{Matching attention}} &
		\multicolumn{1}{l}{\begin{CJK*}{UTF8}{gbsn}\textit{全总(National Federation of Trade Unions)：\fancyhl{100}{全总}(National Federation of Trade Unions)}\end{CJK*}} \\
        
        \multicolumn{1}{l}{\textbf{}} &
		\multicolumn{1}{l}{\begin{CJK*}{UTF8}{gbsn}\textit{等(sort)：\fancyhl{0}{实体、属性、类型}(entity, attributes, kind)\Big|\fancyhl{0}{功能词}(functional word)\Big|\fancyhl{0}{相等}(equal)\Big|\fancyhl{0}{实体、等级}(entity, rank)\Big|\fancyhl{100}{等待}(await)}\end{CJK*}} \\
        
        \multicolumn{1}{l}{\textbf{}} &
		\multicolumn{1}{l}{\begin{CJK*}{UTF8}{gbsn}\textit{单位(workplace)：\fancyhl{8}{单位、量度}(unit, measurement)\Big|\fancyhl{100}{事务、从事、组织}(affairs, engage, organization)}\end{CJK*}} \\
        
        \multicolumn{1}{l}{\textbf{}} &
		\multicolumn{1}{l}{\begin{CJK*}{UTF8}{gbsn}\textit{慰问(convey greetings to)：\fancyhl{100}{安慰、问候}(soothe, sayhello)}\end{CJK*}} \\
        
        \multicolumn{1}{l}{\textbf{}} &
		\multicolumn{1}{p{13cm}}{\begin{CJK*}{UTF8}{gbsn}\textit{本(native)：\fancyhl{100}{主}(primary)\Big|\fancyhl{0}{读物}(readings)\Big|\fancyhl{0}{己}(self)\Big|\fancyhl{0}{事件、实体、根、部件}(event, entity, base, part)\Big|\fancyhl{0}{实体、根、部件}(entity, base, part)\Big|\fancyhl{0}{簿册}(account)\Big|\fancyhl{0}{植物、身、部件}(plant, body, part)\Big|\fancyhl{0.17}{资金、金融}(fund, finance)\Big|\fancyhl{0}{现在}(present)\Big|\fancyhl{0}{特定}(specific)}\end{CJK*}} \\
        
        \multicolumn{1}{l}{\textbf{}} &
		\multicolumn{1}{l}{\begin{CJK*}{UTF8}{gbsn}\textit{教师(teacher)：\fancyhl{100}{人、教、教育、职位}(human, teach, education, occupation)}\end{CJK*}} \\
        
        \multicolumn{1}{l}{\textbf{}} &
		\multicolumn{1}{l}{\begin{CJK*}{UTF8}{gbsn}\textit{市(city)：\fancyhl{100}{地方、市}(place, city)\Big|\fancyhl{75.3}{专、地方、市}(ProperName, place, city)}\end{CJK*}} \\
        
        \multicolumn{1}{l}{\textbf{Sememe-level attention}} &
		\multicolumn{1}{l}{\begin{CJK*}{UTF8}{gbsn}\textit{\fancyhc{0}{全总}\ \fancyhc{0}{等}\ \fancyhc{0.01}{单位}\ \fancyhc{0.25}{慰问}\ \fancyhc{8.5}{本}\ \fancyhc{54.5}{教师}\ \fancyhc{100}{市}}\end{CJK*}} \\
        
        \hline
        
	\end{tabular}
	\caption{One case in the test set. Test Sentence is a negative sentence. The sentence is created by reversing the position of two words of the same POS randomly. The bold words are the words we replaced. Word-level attention, Matching attention, and Sememe-level attention show the dual-attention mechanism visualization during prediction. In Matching attention, the symbol ``\Big|'' separates different senses of the word.}\label{tab_part1_res3}
\end{table*}

\begin{table*}[tb]
	\centering
    \tiny
	\begin{tabular}{@{}l@{}l@{}l@{}}
		\hline
        
        \multicolumn{1}{l}{\textbf{Test Sentence}} &
		\multicolumn{1}{l}{\begin{CJK*}{UTF8}{gbsn}\textit{目前\ ，\ 苏联\ 、\ 美国\ 、\ 加拿大\ 等\ 国\ ，\ 都\ 在\ 竞相\ 研制\ 与\ 改进\ 核动力\ 破冰船}\end{CJK*}}  \\
        
        \multicolumn{1}{l}{} &
        \multicolumn{1}{l}{At present, the Soviet Union, the United States and Canada are competing to develop and improve nuclear-powered icebreakers.}\\
        
        \multicolumn{1}{l}{\textbf{Word-level attention}} &
		\multicolumn{1}{l}{\begin{CJK*}{UTF8}{gbsn}\textit{\fancyhc{0.0794}{目前}\ \fancyhc{4.78}{，}\ \fancyhc{0.0458}{苏联}\ \fancyhc{0.030}{、}\ \fancyhc{0.0345}{美国}\ \fancyhc{0}{、}\ \fancyhc{0.03}{加拿大} \fancyhc{0.072}{等} \fancyhc{0.3}{国} \fancyhc{41}{，} \fancyhc{1.6}{都} \fancyhc{0.2}{在} \fancyhc{0}{竞相} \fancyhc{0.02}{研制} \fancyhc{2.1}{与} \fancyhc{6}{改进} \fancyhc{35.8}{核动力} \fancyhc{100}{破冰船} }\end{CJK*}} \\

        \multicolumn{1}{l}{\textbf{Matching attention}} &
		\multicolumn{1}{l}{\begin{CJK*}{UTF8}{gbsn}\textit{目前(at present)：\fancyhl{100}{时间、现在}(time, present)}\end{CJK*}}   \\
        
        \multicolumn{1}{l}{\textbf{}} &
		\multicolumn{1}{l}{\begin{CJK*}{UTF8}{gbsn}\textit{，：\fancyhl{100}{标点}(punctuation)}\end{CJK*}} \\
        
        \multicolumn{1}{l}{\textbf{}} &
		\multicolumn{1}{p{13cm}}{\begin{CJK*}{UTF8}{gbsn}\textit{苏联(Soviet)：\fancyhl{100}{与特定国家相关、苏联}(relating to country, Soviet Union)\Big|\fancyhl{3.37}{专、国家、地方、政、欧洲}(Proper Name, country, place, politics, Europe)}\end{CJK*}} \\
        
        \multicolumn{1}{l}{\textbf{}} &
		\multicolumn{1}{l}{\begin{CJK*}{UTF8}{gbsn}\textit{、：\fancyhl{100}{标点}(punctuation)}\end{CJK*}} \\
        
        \multicolumn{1}{l}{\textbf{}} &
		\multicolumn{1}{p{13cm}}{\begin{CJK*}{UTF8}{gbsn}\textit{美国(America)：\fancyhl{22.2}{专、北美、国家、地方、政}(Proper Name, North America, country, place, politics)\Big|\fancyhl{100}{与特定国家相关、美国}(relating to country, the US)}\end{CJK*}} \\
        
        \multicolumn{1}{l}{\textbf{}} &
		\multicolumn{1}{l}{\begin{CJK*}{UTF8}{gbsn}\textit{、：\fancyhl{100}{标点}(punctuation)}\end{CJK*}} \\
        
        \multicolumn{1}{l}{\textbf{}} &
		\multicolumn{1}{p{13cm}}{\begin{CJK*}{UTF8}{gbsn}\textit{加拿大(Canada)：\fancyhl{0.8}{专、北美、国家、地方、政}(Proper Name, North America, country, place, politics)\Big|\fancyhl{100}{与特定国家相关、加拿大}(relating to country, Canada)}\end{CJK*}} \\
        
        \multicolumn{1}{l}{\textbf{}} &
		\multicolumn{1}{l}{\begin{CJK*}{UTF8}{gbsn}\textit{等(sort)：\fancyhl{0.8}{实体、属性、类型}(entity, attributes, kind)\Big|\fancyhl{40}{功能词}(functional word)\Big|\fancyhl{0}{相等}(equal)\Big|\fancyhl{0}{实体、等级}(entity, rank)\Big|\fancyhl{100}{等待}(await)}\end{CJK*}} \\

        \multicolumn{1}{l}{\textbf{}} &
		\multicolumn{1}{l}{\begin{CJK*}{UTF8}{gbsn}\textit{国(country)：\fancyhl{100}{国家、地方、政}(country, place, politics)\Big|\fancyhl{34}{姓}\Big|\fancyhl{0.2}{本土}(native)}\end{CJK*}} \\
        
        \multicolumn{1}{l}{\textbf{}} &
		\multicolumn{1}{l}{\begin{CJK*}{UTF8}{gbsn}\textit{，：\fancyhl{100}{标点}(punctuation)}\end{CJK*}} \\
        
        \multicolumn{1}{l}{\textbf{}} &
		\multicolumn{1}{l}{\begin{CJK*}{UTF8}{gbsn}\textit{都(all)：\fancyhl{100}{功能词}(Fuctional word)\Big|\fancyhl{100}{主、地方、市}(primary, place, city)\Big|\fancyhl{100}{国家、国都、地方、政}(country, capital, place, politics)}\end{CJK*}} \\
        
        \multicolumn{1}{l}{\textbf{}} &
		\multicolumn{1}{p{13cm}}{\begin{CJK*}{UTF8}{gbsn}\textit{在(doing)：\fancyhl{100}{依靠}(depend on)\Big|\fancyhl{100}{功能词、进展}(Functional word, going on)\Big|\fancyhl{100}{功能词}(Functional word)\Big|\fancyhl{100}{活着}(alive)\Big|\fancyhl{100}{处于}(situated)\Big|\fancyhl{100}{存在}(exist)} \end{CJK*}} \\
        
        \multicolumn{1}{l}{\textbf{}} &
		\multicolumn{1}{l}{\begin{CJK*}{UTF8}{gbsn}\textit{竞相(compete)：\fancyhl{100}{较量}(have contest)} \end{CJK*}} \\
        
        \multicolumn{1}{l}{\textbf{}} &
		\multicolumn{1}{l}{\begin{CJK*}{UTF8}{gbsn}\textit{研制(develop)：\fancyhl{100}{制造}(produce)} \end{CJK*}} \\
        
        \multicolumn{1}{l}{\textbf{}} &
		\multicolumn{1}{l}{\begin{CJK*}{UTF8}{gbsn}\textit{与(and)：\fancyhl{100}{功能词}(Functional word)} \end{CJK*}} \\
        
        \multicolumn{1}{l}{\textbf{}} &
		\multicolumn{1}{l}{\begin{CJK*}{UTF8}{gbsn}\textit{改进(improve)：\fancyhl{100}{改良}(improve)} \end{CJK*}} \\
        
        \multicolumn{1}{l}{\textbf{}} &
		\multicolumn{1}{l}{\begin{CJK*}{UTF8}{gbsn}\textit{核动力(nuclear power)：\fancyhl{100}{力量、实体}(strength, entity)} \end{CJK*}} \\
        
        \multicolumn{1}{l}{\textbf{}} &
		\multicolumn{1}{l}{\begin{CJK*}{UTF8}{gbsn}\textit{破冰船(icebreaker)：\fancyhl{100}{船}(ship)} \end{CJK*}} \\
        
        \multicolumn{1}{l}{\textbf{Sememe-level attention}} &
		\multicolumn{1}{l}{\begin{CJK*}{UTF8}{gbsn}\textit{\fancyhc{100}{目前}\ \fancyhc{1.01}{，}\ \fancyhc{36}{苏联}\ \fancyhc{2.6}{、}\ \fancyhc{5.74}{美国}\ \fancyhc{37}{、}\ \fancyhc{28.9}{加拿大} \fancyhc{8.4}{等} \fancyhc{1.2}{国} \fancyhc{0.02}{，} \fancyhc{4.7}{都} \fancyhc{9.2}{在} \fancyhc{48}{竞相} \fancyhc{80.4}{研制} \fancyhc{6.6}{与} \fancyhc{10.4}{改进} \fancyhc{33.8}{核动力} \fancyhc{34.7}{破冰船} }\end{CJK*}} \\

        \hline
        
	\end{tabular}
	\caption{One case in the test set. Test Sentence is a positive sentence. Word-level attention, Matching attention, and Sememe-level attention show the dual-attention mechanism visualization during prediction. In Matching attention, the symbol ``\Big|'' separates different senses of the word.}\label{tab_part1_res10}
\end{table*}

\begin{itemize}

\item Without the knowledge in HowNet, the accuracy of the model drops by 1.7\% (in \textbf{w/o HowNet}). The sememe knowledge given by HowNet can provide some fine-grained semantic information, and thus can help the task of SSRD. 

\item It is useful to model the relation between the sentence and HowNet knowledge more properly. We can observe that without the matching mechanism between the sememe-level and the word-level, the accuracy of the model drops by 1.1\% (in \textbf{w/o Match}). It shows matching mechanism can give a more rational and fine-grained semantic representation of the sentence. Furthermore, this sort of representation can help the task of SSRD. 

\item The Dual-attention mechanism is of great help to our task. Without this mechanism, the accuracy of the model drops by 5.4\% (in \textbf{w/o Dual-attention}). It shows this sort of hierarchical attention mechanism in SWM-NN can make use of the information of sentence and HowNet properly to achieve our task.

\item The information from the sentence is very important. Simply averaging the sememe information without any word-level information can make the accuracy of the model drop by 9.6\% (in \textbf{w/o Word-level part}), which is a large margin. The model without word-level part is even worse than some of our baselines. The reason might be that different words have different numbers of sememes in HowNet. The word that has a large number of sememes will get an ambiguous representation under the simple average operation. We cannot properly make use of these words if we do not have the word-level part. 

\item Without the word-level representation of the sentence, the accuracy of the model drops by 0.8\% (in \textbf{w/o Word-level $\bm{c}^w$}). It is a loss that cannot be ignored. Even if we get a proper sememe representation, the representation of sentence in word-level is also helpful in our task.

\end{itemize}

Based on the ablation studies above, every parts in our model is necessary to achieve the best result in the task of SSRD.

\subsection{Case Study}

Here we show some sentences and their dual-attention weight visualization in the test set for case study. Table~\ref{tab_part1_res3} shows a negative sentence that get a correct prediction in our test set. We can see that the ``Word-level attention'' gives higher weights to the word ``\begin{CJK*}{UTF8}{gbsn}\textit{全总}\end{CJK*}''. It might because the word ``\begin{CJK*}{UTF8}{gbsn}\textit{全总}\end{CJK*}'' is the abbreviation for the word ``\begin{CJK*}{UTF8}{gbsn}\textit{全市总工会} (National Federation of Trade Unions)\end{CJK*}'' in Chinese so that it confuses the word-level model. But this sort of situation is not conducive to the prediction. In the ``Matching attention'', we can see that the global attention mechanism weights are mainly correct except the word ``\begin{CJK*}{UTF8}{gbsn}\textit{等} (sort)\end{CJK*}''. After the matching mechanism, we can observe that ``Sememe-level attention'' gives higher weights to the wrong word ``\begin{CJK*}{UTF8}{gbsn}\textit{教师} (teacher)\end{CJK*}'' and ``\begin{CJK*}{UTF8}{gbsn}\textit{市} (city)\end{CJK*}''. This shows that in order to predict correctly, our model gives a higher attention to the wrong words correctly. 
Table~\ref{tab_part1_res10} shows a positive sentence that get a correct prediction in our test set. 
The word ``\begin{CJK*}{UTF8}{gbsn}\textit{破冰船} (icebreaker)\end{CJK*}'' is uncommon in Chinese corpus. We can see that the ``Word-level attention'' gives a higher attention to this word. The sememe knowledge matches its meaning to the word ``\begin{CJK*}{UTF8}{gbsn}\textit{船} (ship)\end{CJK*}''. This word is very common in Chinese corpus. Hence, we can observe that the ``Sememe-level attention'' lowers the attention to the word ``\begin{CJK*}{UTF8}{gbsn}\textit{破冰船} (icebreaker)\end{CJK*}''. These phenomena can be the reason that the model successfully predicts the label of this sentence. From ``Matching attention'', we can observe that most of the word in this sentence get a mainly rational attention weights except the word ``\begin{CJK*}{UTF8}{gbsn}\textit{等} (sort)\end{CJK*}'' and the word ``\begin{CJK*}{UTF8}{gbsn}\textit{都} (all)\end{CJK*}''.

\subsection{Error Analysis}

For error analysis, we first construct four datasets. The permuted sentences in each set are created as follows.

\begin{itemize}

\item \noindent \textbf{Dataset1:}  Replace one word with the same POS randomly.

\item \noindent \textbf{Dataset2:}  Replace two words with the same POS randomly.

\item \noindent \textbf{Dataset3:}  Reverse the position of two words of the same POS randomly.

\item \noindent \textbf{Dataset4:}  Reverse the position of two words randomly.

\end{itemize}

We train our models on each training set and then evaluate on the corresponding test set. Meanwhile we select 500 sentences from each set and ask the human annotators to label. Table~\ref{tab_part1_res4} shows the result of each dataset.

\begin{table}[tb]
	\centering
	\begin{tabular}{l c l c}
		\hline
		\multicolumn{1}{l}{\textbf{Dataset}} &
        \multicolumn{1}{l}{\textbf{Model}} &
        \multicolumn{1}{l}{\textbf{Human}}\\ \hline
        Dataset1 & 35.5\% & 5.0\% \\ 
        Dataset2 & 29.9\%  &  2.2\% \\
        Dataset3 & 32.4\% &  8.6\% \\
        Dataset4 & 28.3\%  &  1.4\% \\ \hline
        
	\end{tabular}
	\caption{Error rate of the model evaluation and the human evaluation for each set.}\label{tab_part1_res4}
\end{table}

From the results shown in Table~\ref{tab_part1_res4}, we can see that 

\begin{itemize} 

\item \noindent  For the model, the most difficult dataset is the dataset1 where the permuted sentences differ in only one word from the true sentences. This shows that the number of words replaced is the biggest challenge for the model. It is partly because that replacing one word with the same POS randomly will exploit polysemy as most of the replaced words have more than one sememes in Hownet. Furthermore, the model is less effective in predicting dataset3, even though the other datasets are replaced by two words. It is partly because that reversing the position of two words of the same POS will swap semantic roles.

\item \noindent  For the human, the most difficult dataset is dataset3. This can also partly show that dataset3 is the most difficult dataset for judging semantic-rationality. As for the other three datasets, both the number of replacement word and the POS of replacement word that affect human judgment. 

\item \noindent  The result of human prediction is much better than that predicted by the model. Among all the dataset, however, the performance of the model on dataset3 is not as bad as the performance of humans on dataset3.

\end{itemize}

\section{Related Work}
As more and more sentences are generated automatically, it is important to measure the quality of the sentences. Traditionally evaluation involves human judgments of different quality metrics~\cite{mani2001automatic}. It is very expensive and difficult to conduct on a frequent basis. There exist some automatic methods to evaluate the quality of the sentence. Some are based on language models, some are based on the similarity with human-written sentences and some are the machine learning models based on statistical features.

There are a wide variety of different language modeling and smoothing techniques. Such as Good-Turing discounting~\cite{nadas1984estimation,church1991comparison}, Witten-Bell discounting~\cite{witten1991zero}, and varieties of class-based Ngram models that used information about word classes. One of the most commonly used and best performing $N$-gram smoothing methods is the interpolated Kneser-Ney algorithm~\cite{kneser1995improved,chen1999empirical}. The drawback of language models lies in that these models only give the common sentences high scores and the infrequent sentences always get low scores even if they are rational. For similarity based methods, there exist some automatic metrics such as BLEU ~\cite{papineni2002bleu}, ROUGE~\cite{lin2004rouge}, SARI~\cite{xu2016optimizing}. These metrics count the $N$-gram matching numbers between the generated sentences and the reference sentences. Then these metrics use these numbers to compute a score. However, the main problem lies in that in many NLP tasks, human-written references are not available, which leads to the failure of this method. For some statistical feature-based methods such as decision tree~\cite{eneva2001learning}, they only use the statistical information of the sentence. However, it is also essential to use the semantic information of the sentences when evaluating their quality. Proper use of semantic information can help improve NLP performance, ~\newcite{ma2017improving} improve semantic relevance for sequence-to-sequence learning of Chinese social media text summarization. ~\newcite{ma2017semantic} proposed a novel semantic relevance based neural network for text summarization and text simplification.
 
The knowledge base can help the model to understand the semantic meaning of a sentence better. ~\newcite{zhang2018duplicate} identify duplicate question by integrating knowledge base FrameNet with neural networks. ~\newcite{shi2016knowledge} use knowledge-based semantic embedding in machine translation. One of the most well-known semantic knowledge bases is HowNet ~\cite{zhendong2006hownet}, which is a sememe knowledge base. HowNet has been widely used in various Chinese NLP tasks, such as word sense disambiguation ~\cite{duan2007word}, named entity recognition ~\cite{10.1007/978-3-319-50496-4_38}, word representation ~\cite{niu2017improved},~\newcite{zeng2018chinese} proposed to expand the Linguistic Inquiry and Word Count ~\cite{pennebaker2001linguistic} lexicons based on word sememes, and ~\newcite{li2018sememe} proposed a task called sememe prediction to learn semantic knowledge from unstructured textual Wiki descriptions.

To overcome the disadvantages of the methods mentioned above, we address the task of automatic semantic rationality evaluation by using the semantic information expressed by sememes. We use a novel model to combine the sentence and the sememe knowledge base HowNet to deal the task.

\section{Conclusion}
In this paper, we propose the task of sentence semantic rationality detection (SSRD), which aims to identify whether the sentence is rational in semantics. To deal with the difficulties in this task and overcome the disadvantages of current methods, we propose a Sememe-Word-Matching Neural Network model that not only considers the information of the sentences, but also makes use of the sememe information in knowledge base HowNet. Furthermore, our model selects the proper sememe information by the matching mechanism. Experimental results show that our model can outperform various baselines by a large margin. 

Further experiments show that although our model has achieved promising results, there is still a big gap compared with the artificial results. How to make better use of other knowledge bases in this sort of task will be our future work.

\bibliography{emnlp2018}

\begin{thebibliography}{29}
\expandafter\ifx\csname natexlab\endcsname\relax\def\natexlab#1{#1}\fi

\bibitem[{Bahdanau et~al.(2014)Bahdanau, Cho, and Bengio}]{bahdanau2014neural}
Dzmitry Bahdanau, Kyunghyun Cho, and Yoshua Bengio. 2014.
\newblock Neural machine translation by jointly learning to align and
  translate.
\newblock \emph{arXiv preprint arXiv:1409.0473}.

\bibitem[{Bloomfield(1926)}]{bloomfield1926set}
Leonard Bloomfield. 1926.
\newblock A set of postulates for the science of language.
\newblock \emph{Language}, 2(3):153--164.

\bibitem[{Chen and Goodman(1999)}]{chen1999empirical}
Stanley~F Chen and Joshua Goodman. 1999.
\newblock An empirical study of smoothing techniques for language modeling.
\newblock \emph{Computer Speech \& Language}, 13(4):359--394.

\bibitem[{Chomsky(1956)}]{chomsky1956three}
Noam Chomsky. 1956.
\newblock Three models for the description of language.
\newblock \emph{IRE Transactions on information theory}, 2(3):113--124.

\bibitem[{Church and Gale(1991)}]{church1991comparison}
Kenneth~W Church and William~A Gale. 1991.
\newblock A comparison of the enhanced good-turing and deleted estimation
  methods for estimating probabilities of english bigrams.
\newblock \emph{Computer Speech \& Language}, 5(1):19--54.

\bibitem[{Dong and Dong(2006)}]{zhendong2006hownet}
Zhendong Dong and Qiang Dong. 2006.
\newblock \emph{Hownet And The Computation Of Meaning (With Cd-rom)}.
\newblock World Scientific.

\bibitem[{Duan et~al.(2007)Duan, Zhao, and Xu}]{duan2007word}
Xiangyu Duan, Jun Zhao, and Bo~Xu. 2007.
\newblock Word sense disambiguation through sememe labeling.
\newblock In \emph{IJCAI}, pages 1594--1599.

\bibitem[{Eneva et~al.(2001)Eneva, Hoberman, and Lita}]{eneva2001learning}
Elena Eneva, Rose Hoberman, and Lucian Lita. 2001.
\newblock Learning within-sentence semantic coherence.
\newblock In \emph{Proceedings of the 2001 Conference on Empirical Methods in
  Natural Language Processing}.

\bibitem[{Hochreiter and Schmidhuber(1997)}]{Hochreiter1997Long}
Sepp Hochreiter and J{\"u}rgen Schmidhuber. 1997.
\newblock Long short-term memory.
\newblock \emph{Neural computation}, 9(8):1735--1780.

\bibitem[{Kim(2014)}]{cnn}
Yoon Kim. 2014.
\newblock Convolutional neural networks for sentence classification.
\newblock In \emph{Proceedings of the 2014 Conference on Empirical Methods in
  Natural Language Processing, {EMNLP} 2014, October 25-29, 2014, Doha, Qatar,
  {A} meeting of SIGDAT, a Special Interest Group of the {ACL}}, pages
  1746--1751.

\bibitem[{Kingma and Ba(2014)}]{KingmaBa2014}
Diederik~P. Kingma and Jimmy Ba. 2014.
\newblock Adam: {A} method for stochastic optimization.
\newblock \emph{CoRR}, abs/1412.6980.

\bibitem[{Kneser and Ney(1995)}]{kneser1995improved}
Reinhard Kneser and Hermann Ney. 1995.
\newblock Improved backing-off for m-gram language modeling.
\newblock In \emph{Acoustics, Speech, and Signal Processing, 1995. ICASSP-95.,
  1995 International Conference on}, volume~1, pages 181--184. IEEE.

\bibitem[{Li et~al.(2018)Li, Ren, Dai, Wu, Wang, and Sun}]{li2018sememe}
Wei Li, Xuancheng Ren, Damai Dai, Yunfang Wu, Houfeng Wang, and Xu~Sun. 2018.
\newblock Sememe prediction: Learning semantic knowledge from unstructured
  textual wiki descriptions.
\newblock \emph{arXiv preprint arXiv:1808.05437}.

\bibitem[{Li et~al.(2016)Li, Wu, and Lv}]{10.1007/978-3-319-50496-4_38}
Wei Li, Yunfang Wu, and Xueqiang Lv. 2016.
\newblock Improving word vector with prior knowledge in semantic dictionary.
\newblock In \emph{Natural Language Understanding and Intelligent
  Applications}, pages 461--469, Cham. Springer International Publishing.

\bibitem[{Lin(2004)}]{lin2004rouge}
Chin-Yew Lin. 2004.
\newblock Rouge: A package for automatic evaluation of summaries.
\newblock \emph{Text Summarization Branches Out}.

\bibitem[{Ma and Sun(2017)}]{ma2017semantic}
Shuming Ma and Xu~Sun. 2017.
\newblock A semantic relevance based neural network for text summarization and
  text simplification.
\newblock \emph{arXiv preprint arXiv:1710.02318}.

\bibitem[{Ma et~al.(2017)Ma, Sun, Xu, Wang, Li, and Su}]{ma2017improving}
Shuming Ma, Xu~Sun, Jingjing Xu, Houfeng Wang, Wenjie Li, and Qi~Su. 2017.
\newblock Improving semantic relevance for sequence-to-sequence learning of
  chinese social media text summarization.
\newblock \emph{arXiv preprint arXiv:1706.02459}.

\bibitem[{Mani(2001)}]{mani2001automatic}
Inderjeet Mani. 2001.
\newblock \emph{Automatic summarization}, volume~3.
\newblock John Benjamins Publishing.

\bibitem[{Nadas(1984)}]{nadas1984estimation}
Arthur Nadas. 1984.
\newblock Estimation of probabilities in the language model of the ibm speech
  recognition system.
\newblock \emph{IEEE Transactions on Acoustics, Speech, and Signal Processing},
  32(4):859--861.

\bibitem[{Niu et~al.(2017)Niu, Xie, Liu, and Sun}]{niu2017improved}
Yilin Niu, Ruobing Xie, Zhiyuan Liu, and Maosong Sun. 2017.
\newblock Improved word representation learning with sememes.
\newblock In \emph{Proceedings of the 55th Annual Meeting of the Association
  for Computational Linguistics (Volume 1: Long Papers)}, volume~1, pages
  2049--2058.

\bibitem[{Papineni et~al.(2002)Papineni, Roukos, Ward, and
  Zhu}]{papineni2002bleu}
Kishore Papineni, Salim Roukos, Todd Ward, and Wei-Jing Zhu. 2002.
\newblock Bleu: a method for automatic evaluation of machine translation.
\newblock In \emph{Proceedings of the 40th annual meeting on association for
  computational linguistics}, pages 311--318. Association for Computational
  Linguistics.

\bibitem[{Pascanu et~al.(2013)Pascanu, Mikolov, and Bengio}]{gradientclip}
Razvan Pascanu, Tomas Mikolov, and Yoshua Bengio. 2013.
\newblock On the difficulty of training recurrent neural networks.
\newblock In \emph{International Conference on Machine Learning}, pages
  1310--1318.

\bibitem[{Pennebaker et~al.(2001)Pennebaker, Francis, and
  Booth}]{pennebaker2001linguistic}
James~W Pennebaker, Martha~E Francis, and Roger~J Booth. 2001.
\newblock Linguistic inquiry and word count: Liwc 2001.
\newblock \emph{Mahway: Lawrence Erlbaum Associates}, 71(2001):2001.

\bibitem[{Shi et~al.(2016)Shi, Liu, Ren, Feng, Li, Zhou, Sun, and
  Wang}]{shi2016knowledge}
Chen Shi, Shujie Liu, Shuo Ren, Shi Feng, Mu~Li, Ming Zhou, Xu~Sun, and Houfeng
  Wang. 2016.
\newblock Knowledge-based semantic embedding for machine translation.
\newblock In \emph{Proceedings of the 54th Annual Meeting of the Association
  for Computational Linguistics (Volume 1: Long Papers)}, volume~1, pages
  2245--2254.

\bibitem[{Srivastava et~al.(2014)Srivastava, Hinton, Krizhevsky, Sutskever, and
  Salakhutdinov}]{dropout}
Nitish Srivastava, Geoffrey~E. Hinton, Alex Krizhevsky, Ilya Sutskever, and
  Ruslan Salakhutdinov. 2014.
\newblock Dropout: a simple way to prevent neural networks from overfitting.
\newblock \emph{Journal of Machine Learning Research}, 15(1):1929--1958.

\bibitem[{Witten and Bell(1991)}]{witten1991zero}
Ian~H Witten and Timothy~C Bell. 1991.
\newblock The zero-frequency problem: Estimating the probabilities of novel
  events in adaptive text compression.
\newblock \emph{Ieee transactions on information theory}, 37(4):1085--1094.

\bibitem[{Xu et~al.(2016)Xu, Napoles, Pavlick, Chen, and
  Callison-Burch}]{xu2016optimizing}
Wei Xu, Courtney Napoles, Ellie Pavlick, Quanze Chen, and Chris Callison-Burch.
  2016.
\newblock Optimizing statistical machine translation for text simplification.
\newblock \emph{Transactions of the Association for Computational Linguistics},
  4:401--415.

\bibitem[{Zeng et~al.(2018)Zeng, Yang, Tu, Liu, and Sun}]{zeng2018chinese}
Xiangkai Zeng, Cheng Yang, Cunchao Tu, Zhiyuan Liu, and Maosong Sun. 2018.
\newblock Chinese liwc lexicon expansion via hierarchical classification of
  word embeddings with sememe attention.

\bibitem[{Zhang et~al.(2018)Zhang, Sun, and Wang}]{zhang2018duplicate}
Xiaodong Zhang, Xu~Sun, and Houfeng Wang. 2018.
\newblock Duplicate question identification by integrating framenet with neural
  networks.
\newblock In \emph{AAAI 2018}.

\end{thebibliography}
\bibliographystyle{acl_natbib_nourl}

\end{document}